\documentclass[conference]{IEEEtran}

\usepackage{microtype}
\usepackage{graphicx}
\usepackage{subfigure}
\usepackage{booktabs} 
\usepackage[square,numbers]{natbib}

\usepackage[utf8]{inputenc} 
\usepackage[T1]{fontenc}    
\usepackage{hyperref}       
\usepackage{url}            
\usepackage{booktabs}       
\usepackage{amsfonts}       
\usepackage{nicefrac}       
\usepackage{microtype}      

\usepackage{graphicx}
\usepackage{amsmath,amssymb,amsfonts}
\usepackage{multicol} 
\usepackage{caption} 
\usepackage{wrapfig}
\usepackage[capitalise]{cleveref}
\usepackage[colorinlistoftodos,prependcaption]{todonotes}
\usepackage{balance} 

\def\ctrl{a}

\def\real{\mathbb R}
\def\natural{\mathbb N}

\def\comment#1{}

\def\NULL{\text{NULL}}

\def\EOS{\text{EOS}}

\def\READ{\text{READ} }
\def\WRITE{\text{WRITE} } 

\def\fin{\text{fin}}

\def\eqref#1{(\ref{#1})}
\def\Beq#1\Eeq{\begin{equation}#1\end{equation}}
\def\Beqo#1\Eeqo{\begin{equation*}#1\end{equation*}}
\def\Beqs#1\Eeqs{\begin{align}#1\end{align}}
\def\Beqso#1\Eeqso{\begin{align*}#1\end{align*}}

\begin{document}

\title{Reinforcement Learning for  \\on-line Sequence Transformation} 


\author{\IEEEauthorblockN{Grzegorz Rype\'s\'c, \L{}ukasz Lepak, Pawe\l{} Wawrzy\'nski\\
 Warsaw University of Technology, Institute of Computer Science, Warsaw, Poland}}

\maketitle

\begin{abstract}
In simultaneous machine translation (SMT), an output sequence should be produced as soon as possible, without reading the whole input sequence.
This requirement creates a trade-off between translation delay and quality because less context may be known during translation. 
In most SMT methods, this trade-off is controlled with parameters whose values need to be tuned.
In this paper, we introduce an SMT system that learns with reinforcement and is able to find the optimal delay in training. 
We conduct experiments on Tatoeba and IWSLT2014 datasets against state-of-the-art translation architectures.
Our method achieves comparable results on the former dataset, with better results on long sentences and worse but comparable results on the latter dataset. 
\end{abstract}

\section{Introduction} 

Simultaneous machine translation (SMT) can be defined as producing output sequence tokens while reading input sequence tokens in an on-line fashion.
These tokens may represent words in given languages, chunks of audio streams, or any other sequential data.
The main difference between SMT and more general neural machine translation (NMT) is how the input and output sequences are processed.
Most NMT methods read all input tokens and then generate the output sequence.
Because of this, even though efficient state-of-the-art NMT methods exist, they cannot be used in SMT applications.
Also, SMT methods need to consider the trade-off between delay and quality, as faster translation implies less context from the input. 
In most cases, this trade-off has to be optimized by checking various parameter settings, which is resource- and time-consuming.

SMT can be decomposed into a~sequence of readings of the input tokens and writings of the output tokens. Reinforcement learning (RL) \cite{2018sutton+1} is often applied to train SMT systems that sequentially choose between these actions and/or choose the written token. 
In this paper, we present an RL-based method with self-learning delay. Unlike in other approaches, we apply bootstrapping in training, which means that the sequences translated can be in principle infinite. We conduct experiments on Tatoeba and IWSLT2014 datasets against state-of-the-art translation architectures.
Our method achieves comparable results on the former dataset, with better results on long sentences and worse but comparable results on the latter dataset. 

The paper is organized as follows. 
Section~\ref{sec:related-work} overviews literature related to neural machine translation, reinforcement learning, and simultaneous machine translation. 
Section~\ref{sec:prob:def} formally defines the problem considered in this paper. 
Section~\ref{sec:method} presents our method. 
Section~\ref{sec:experiments} describes simulations evaluating the presented architecture. 
Section~\ref{sec:discussion} discusses the experimental results and limitations of our approach. 
Section~\ref{sec:conclusions} concludes the paper.

\section{Related work} 
\label{sec:related-work} 

\paragraph{Neural machine translation (NMT).}
A basic architecture for neural machine translation includes an encoder that is fed with the input sequence; its final state becomes the initial state of a~decoder that produces the output sequence \cite{2014sutskever+2}. In order to produce the right output, attention must be paid to significant input tokens. Attention was introduced to the encoder-decoder architecture in \cite{2015bahdanau+2} and \cite{2015luong+2}. 
An architecture for NMT that is based solely on attention is Transformer \cite{2017vaswani+7}. Recurrent neural networks (RNN) were applied to capture short-term dependencies in input sequences and combined with multilayer attention in R-Transformer \cite{2019wang+3}. However, all these architectures only produce output when given the whole input sequence and hence are not applicable to on-line translation. 

\paragraph{Reinforcement learning (RL).} 
RL is a general framework for adaptation in the context of sequential decision making under uncertainty \cite{2018sutton+1}. In this framework, an agent operates in discrete time, at each instant observing the~state of its environment and taking action. Subsequently, the environment state changes, and the agent receives a~numeric reward. Both the next state and the reward result from the previous state and action. By repeatedly facing the~sequential decision problem in the same environment, the agent learns to designate actions in current environment states to be able to expect the highest future rewards.  

In the context of this paper, especially interesting is the case where the agent cannot observe its state but only the~value of a~certain function of the state. This case is modeled as the Partially Observable Markov Decision Process (POMDP) \cite{2019kapturowski+4}. In this model, the agent needs to collect subsequent observations to be able to recognize its current situation at any specific time. This can be done effectively with an RNN. Deep Recurrent Q-Learning \cite{2015hausknecht+1} is an RL method for POMDP, which applies an RNN for that purpose. 

RL has been applied to neural machine translation to optimize a~policy, which, given an input sentence, assigned maximum probability to the corresponding output sentence. RL was applied this way to optimize the translation quality expressed in BLEU \cite{wu2018study} which is not directly differentiable. 
RL has also been applied to train a~random generator of sentences in a generative adversarial architecture~\cite{2017yu+3}. A~similar architecture has been applied for the sequential generation of graphs \cite{2017guimaraes+4}.

\paragraph{Simultaneous Machine Translation (SMT).} 
A number of SMT methods use reinforcement learning.
One of the first examples of using RL for SMT was presented in \cite{2014grissom+4}. 
It uses imitation learning from the optimal sequence of actions to learn a policy for the system. 
\citet{2017gu+3} introduced a two-action framework, where the agent can read an input token, named READ, or write a new output token, named WRITE.
This framework serves as a baseline for many new SMT methods, with authors extending and modifying it to achieve better results. 
The proposed reward function is based on the achieved BLEU score \cite{2002papineni+3} and the translation delay metrics proposed by the authors, with the trade-off between delay and translation quality controlled by setting appropriate parameter values. 
In \citep{2018alinejad+2}, a third action was added, named PREDICT, which works similarly to READ, but instead of reading an input token, it predicts this token.
The reward function was also changed to include predictions' quality, with delay-quality trade-off still controlled by parameters. 
In \citep{2018dalvi+3}, a commonly used NMT encoder-decoder structure was modified to work with SMT by making encoder, and attention dynamically change after every READ and adding an incremental decoder, which outputs a token from them after every WRITE. 
In \citep{2019zheng+3}, a method was proposed for extracting action sequences from NMT architectures, which were later used with sentence pairs in imitation learning to learn an optimal policy. 
Recently, reinforcement learning was used in multimodal translation \cite{2021ive+5}, utilizing text and visual data to improve the quality of translations. 

Not every SMT method uses reinforcement learning.
In \cite{2019ma+11}, the ''wait-$k$'' strategy was proposed, which produces a new output token with a fixed delay equal to $k$.
It can be easily implemented in commonly used NMT architectures, shown by modifying the original Transformer. 
In \citep{2020ren+6}, the ''wait-k'' strategy was used in speech-to-text task, showing it is efficient in applications other than machine translation. 

\section{Problem definition} 
\label{sec:prob:def} 

We consider input sequences, $x = (x_i)_{i=0}^{|x|-1}$, that contain tokens, $x_i \in \real^d$, $d\in\natural$. The input sequences correspond to target sequences, $y = (y_j)_{j=0}^{|y|-1}$, $y_j \in \real^{d'}$, $d'\in\natural$. The sequences are of variable lengths presented by the $|\cdot|$ function. An {\it interpreter agent} is fed with subsequent tokens from $x$ and produces tokens of an~output sequence, $(z_j)_{j=0}^{|z|-1}$, $z_j \in \real^{d'}$ on the basis of $x$. 

Three special tokens playing various roles exist in both the input and the output space. They are: 
\begin{itemize} 
\item NULL --- a missing element,
\item EOS --- denotes the last element of each sequence, 
\item PAD --- an element concatenated to sequences after EOS for technical reasons. 
\end{itemize} 
For brevity, we will assume $x_i\!=\!\text{PAD}, y_j\!=\!\text{PAD}, z_j\!=\!\text{PAD}$ for, respectively, $i\geq|x|, j\geq|y|, j\geq|z|$. 

Given $x$, the agent should produce $z$ that minimizes the~quality index in the form 
\Beq \label{quality:idx} 
    J(y,z) = \sum_{j=0}^{K-1} L(y_j,z_j).
\Eeq
The loss $L$ penalizes mistranslation;  $L(\text{PAD},\text{PAD})=0$; $K$ is a~number larger than any~$|y|$. The sequence $z$ that minimizes \eqref{quality:idx} is of length $|y|$, contains tokens equal to those in $y$, and ends with EOS.   

We also require the interpreter agent to be of limited capacity but handle sequences of arbitrarily large lengths. In other words, we require the agent to operate on-line, i.e., it is fed with subsequent tokens of the input sequence and simultaneously produces subsequent tokens of the output sequence. 

\section{Method} 
\label{sec:method} 

\subsection{Reinforcement learning to transform sequences} 

We formalize the transformation of one sequence into another as an~iterative decision process. At each of its instants, an agent reads a~subsequent token from the input sequence or writes a~subsequent token of the output sequence, similarly to \cite{2017gu+3}. That is, at each instant, the agent executes one of two actions: 
\begin{itemize} 
\item READ --- another input token is read. This action is useful when it is (still) unclear what output token should be produced. 
\item WRITE --- a~subsequent output token is produced. This action is useful when a~certain comprehensive portion of input tokens have been read, and a~subsequent part of its interpretation can be presented. 
\end{itemize} 
A {\it policy} is a method of selecting actions and producing output tokens based on tokens read and those produced so far. 

After execution of some of the actions, the agent receives numerical {\it rewards}. Let the rewards received during the process be denoted by $r = (r_k)_{k=0}^{|r|-1}$. A~reward, $r_k$, is emitted at the following times:
\begin{itemize} 
\item 
An output token, $z_j$, has just been written. Then $r_k$ is the negative cost of mistranslation, i.e. 
\Beq
    r_k = -L(y_j,z_j). 
\Eeq
\item 
A whole input sequence has been read, and the \READ action is taken. This action does not make sense at this time. Therefore, for a certain constant $M>0$, we have  
\Beq
    r_k = -M. 
\Eeq
\end{itemize} 

Let $n(t)$ be the number of rewards emitted before the $t$-th action. The quality criterion for the policy is maximization of future discounted rewards. That is, at each time $t$ the expected value of the {\it return} 
\Beq \label{def:Ret} 
    R_t = \sum_{k=n(t)}^{|r|-1} \gamma^{k-n(t)} r_k
\Eeq
should be maximized, where $\gamma\in(0,1)$ is a~discount factor. 
In one episode of its operation, the agent transforms a~single sequence. It stops producing additional output tokens when it has outputted \EOS. 

In training, the agent, not having learned how to finish sequences, must be prevented from producing them infinitely long. Here we assume that an episode of training is terminated when the agent has produced as many tokens as in the target sequence $y$. The last target token is EOS, which is enough for the agent to learn to finish the output sequences. 

Usually, in reinforcement learning \cite{2018sutton+1} a~reward comes after each action. However, here we want the agent to be rewarded only for the tokens it produces, bearing in mind that it does not produce them with READ actions. Rewards equal to zero for such actions do not make sense here because they could encourage the agent to maximize the sum of discounted rewards by postponing the production of output. Therefore, here we admit actions that are not immediately followed by rewards. Those emitted rewards have their own indices and are discounted according to them. 

At each instant of its operation, a~{\it state} of the agent's environment consists of the tokens the agent has read so far and the tokens it has written so far. However, before taking another action, it is only fed with the next input token and with the last written token. Therefore, the agent's environment is partially observable. 

\subsection{Architecture} 

\begin{figure}
    \centering
    \includegraphics[width=0.6\linewidth]{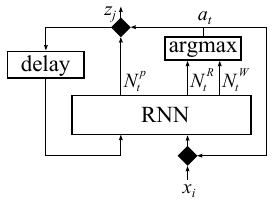}
    \caption{Proposed architecture for on-line sequence transformation. The black squares represent passing/delaying $x_i$ and outputting/skipping $N^p_t$ depending on the action $\ctrl_t$. }
    \label{fig:RLST-structure}
\end{figure}

We propose an~architecture that learns to make the actions discussed above. The policy has the form of a recurrent neural network. Its input size is $d+d'$. In an instant of its operation it is fed with a~subsequent input token concatenated with a~preceding output token. Specifically, the first input to the network is the pair $(x_0,\NULL)$. Let us assume that the agent has already read $i$ input tokens and produced $j$ output tokens. Thus, after the READ action, the network input is $(x_i,\NULL)$. After the WRITE action, the network input is $(\NULL,z_{j-1})$. 

In training {\it teacher forcing} can also be applied: The agent is fed not with the tokens it has already outputted but with target tokens. 

Output of the network is of size $d'+2$. The network produces a~$d'$-dimensional {\it potential output token} and $2$ scalar {\it return estimates} that approximate returns \eqref{def:Ret} expected if actions WRITE and READ, respectively, were taken. 

Let $N_t$ be the $d'+2$-dimensional output of the network at $t$-th instant. It is composed as  
$$
    N_t = [N^{p}_t,N^{W}_t,N^R_t], 
$$
where $N^{p}_t \in \real^{d'}$ is the potential output token, and $N^{W}_t, N^R_t \in \real$ are the return estimates for the WRITE and READ actions, respectively. The architecture is depicted in Fig.~\ref{fig:RLST-structure}. 

The network output that estimates the return corresponding to the just taken action $a_t$ is trained to approximate the conditional expected value  
\Beq \label{Q_t} 
    Q_t(a_t) = E(R_t|C_t), 
\Eeq
where the condition $C_t$ includes the following: 
\begin{enumerate} 
\item 
The action just taken is $a_t$.
\item 
Subsequently, those actions are selected, which correspond to the network return estimates with maximum values. 
\item 
Input tokens read so far, and output tokens produced so far. At the time $t$, the rest of the tokens are unknown, thereby remaining random vectors. 
\end{enumerate} 
The actions actually taken are usually selected as those maximizing the return estimates given by the network. However, with a~small probability, the agent chooses the other action since it needs to explore different actions to learn their consequences. Therefore, we will not estimate $Q_t(a_t)$ based on the actual return, but on a~recursion instead. Specifically, let us denote by $j$ the number of output tokens produced before the analyzed action is taken. A~simple analysis reveals that $Q_t(a_t)$ \eqref{Q_t} satisfies the following recursive equation: 
\Beq \label{Q_t:recursive} 
    Q_t(a_t) = E\left\{\left.\begin{array}{l} 
    -M  + \gamma\max_b Q_{t+1}(b) \\
    \quad \text{if } a_t=\READ, \fin(x) \\ 
    \max_b Q_{t+1}(b) \\
    \quad \text{if }a_t=\READ, \lnot\fin(x) \\ 
    -L(y_j,z_j) + \gamma\max_b Q_{t+1}(b) \\
    \quad \text{if } a_t=\WRITE, j<|y|-1 \\ 
    -L(y_j,z_j) \\
    \quad \text{if } a_t=\WRITE, j=|y|-1     \end{array}\right|a_t\right\} 
\Eeq
where $\fin(x)$ means that all $x$ tokens have been read. 
The condition for the above expectation is that the action actually taken is~$a_t$. 

Target values for the network will be based on the above recursive equation and the fact that $Q_{t+1}(\READ)$ and $Q_{t+1}(\WRITE)$ are estimated by $N^R_{t+1}$ and $N^{W}_{t+1}$, respectively. Therefore, the network outputs at time $t$ are trained as follows.\footnote{We apply the notation: 
$$
    [predicate] = \left\{\begin{array}{l l} 1 & \text{ if } predicate \text{ is true} \\ 
    0 & \text{ otherwise.}\end{array}\right.
$$} 
After the READ action, when $x$ is not finished yet, $N^R_t$ is adjusted: 
\Beq 
    N^R_t \leftarrow \max\{N^R_{t+1},N^{W}_{t+1}\}. 
\Eeq
After the READ action, when $x$ is already finished: 
\Beq
    N^R_t \leftarrow - M + \gamma\max\{N^R_{t+1},N^{W}_{t+1}\}. 
\Eeq
After a WRITE action, the return estimate for the WRITE action is adjusted as 
\Beq
    N^{W}_t \leftarrow - L(N^{p},z_j)  
    + [j<|y|-1]\gamma\max\{N^R_{t+1},N^W_{t+1}\}, \\
\Eeq 
Also, the potential output is adjusted 
\Beq
    N^{p}_t \leftarrow y_j. 
\Eeq

\subsection{Weighting losses due to mistranslation and return estimation}

The network produces outputs of two qualitatively different kinds: the potential output tokens and the return estimates. The network training requires minimization of an~aggregated loss that combines a~loss due to mistranslation and a~loss due to return estimation. We propose to normalize these losses with their averages defined below. 

Let $n=1,2,\dots$ be a training minibatch index. We average original mistranslation losses, $L^M_n$, and original estimation losses, $L^E_n$, according to 
\Beqs
    \bar L^M_n & = w_n \bar L^M_{n-1} + (1-w_n) L^M_n, \label{barLM}\\
    \bar L^E_n & = w_n \bar L^E_{n-1} + (1-w_n) L^E_n \label{barLE} 
\Eeqs
where $\bar L^M_0 = \bar L^E_0 =0$, and 
\Beq
    w_n = 
	\rho (1-\rho^{n-1})/(1-\rho^n), 
\Eeq
where $\rho \in (0,1)$ is the decay factor, e.g.~$\rho=0.99$. The terms (\ref{barLM},\ref{barLE}) approximate arithmetic means for small~$n$, and exponential moving average for larger~$n$. 

Training the network aims at minimizing the aggregated loss in the form 
\Beq \label{def:L_n} 
    L_n = L^M_n / \bar L^M_n + \eta(n,n_0) L^E_n / \bar L^E_n. 
\Eeq
The term $\eta(n,n_0)$ is a relative weight of the estimation loss for the current minibatch/epoch index $n$ and the (expected) total number $n_0$ of minibatches/epochs in the whole training. For small $n$ this weight should be small: $\eta(n,n_0) \approx \eta_{min}$, since high accuracy of future rewards is pointless when quality of outputted tokens is poor. $\eta(n,n_0)$ is gradually growing with $n$ to a~certain asymptote, $\eta_{max}$. $\eta_{min}$ and $\eta_{max}$ are hyperparameters of the training process, e.g. 1/50 and 1/5, respectively. The $\eta$ function may have the form 
\Beq \label{eta(n,n0)} 
    \eta(n,n_0) = \eta_{max} - (\eta_{max}-\eta_{min})\exp(-3n/n_0). 
\Eeq

\section{Experimental study} 
\label{sec:experiments} 

In this section, we demonstrate the effectiveness of our proposed architecture, henceforth called RLST (Reinforcement Learning for on-line Sequence Transformation). We perform experiments with seven machine translation tasks. They are based on datasets taken from Tatoeba \cite{2020tatoeba} and dataset taken from IWSLT2014 \cite{ott2019fairseq}. 

In our machine translation tasks, the input sequence consists of tokens representing words of a sentence in a source language. The aim is to generate a~sequence of tokens with the same sentence meaning as the source sequence. We conduct experiments on datasets presented in Table \ref{tab:nmt_datasets} which contains basic statistics on the source and target languages datasets, sizes of source and target dictionaries, and numbers of sentences in each data split. For Tatoeba datasets, we also separate long test splits, where source sentences have more than 22 tokens. The long test split allows us to compare how models deal with longer input sentences. For all datasets, we compare our proposed RLST architecture with state-of-the-art machine translation architectures, namely encoder-decoder with attention \cite{2015bahdanau+2} and Transformer \cite{2017vaswani+7}. For both encoder-decoder and Transformer, the minimized loss is cross-entropy. For RLST, we quantify $L^M_n$ and $L^E_n$ in \eqref{def:L_n} as cross-entropy and mean square error, respectively. 

In our experiments, we employed the following procedure to optimize hyperparameters of the compared architectures. For Tatoeba datasets, we optimized the hyperparameters manually for all three architectures to obtain their best BLEU score \cite{2002papineni+3} on the En-Es language pair and applied these values to all language pairs. For IWSLT2014 datasets, we optimized the hyperparameters of RLST manually and took the hyperparameters for the Transformer from \cite{2017vaswani+7} and for the encoder-decoder with attention from \cite{wiseman2016sequence}. 

Our simulation experiments have been performed on a PC equipped with AMD Ryzen\texttrademark Thread\-ripper\texttrademark 1920X, 64GB RAM, 4×NVIDIA\texttrademark GeForce\texttrademark RTX 2070 Super\texttrademark. 

\begin{table*}
    \centering
    \begin{tabular}{l|l|r|r|r|r|r|r}
         Dataset & Abbr & Src. & Trg. & Train & Valid. & Test & Long \\
         & & dict. & dict. & set & set & set & test \\
         \hline
         Tatoeba Spanish-English & Tat Es-En & 13 288 & 8 960 & 124 179 & 41 393 & 41 394 & 2 387 \\
         Tatoeba French-English & Tat Fr-En & 13 792 & 10 056 & 161 283 & 53 761 & 53 762 & 2 613 \\
         Tatoeba English-Spanish & Tat En-Es & 8 690 & 12 698 & 115 026 & 38 342 & 38 342 & 2 325 \\
         Tatoeba English-Russian & Tat En-Ru & 10 009 & 21 820 & 241 785 & 80 595 & 80 595 & 1 756 \\
         Tatoeba English-German & Tat En-De & 10 504 & 15 276 & 170 347 & 56 782 & 56 783 & 3 805 \\
         IWSLT2014- & IWSLT- &&&&&& \\ 
         German-English & De-En & 8 848 & 6 632 & 160 239 & 7 283 & 6 750 & --- \\
         IWSLT2014- & IWSLT- &&&&&& \\ 
         English-German & En-De & 6 632 & 8 848 & 160 239 & 7 283 & 6 750 & --- \\
    \end{tabular}
    \caption{Basic statistics of machine translation datasets.}
    \label{tab:nmt_datasets}
\end{table*}

\subsection{Tatoeba}

Tatoeba datasets \cite{2020tatoeba} contain various, mostly unrelated, sentences and their translations provided by the community. We preprocess them using spaCy tokenizer \cite{spacy} and replace tokens that appear in training corpora less than three times with a unique token representing an unknown word. We also remove duplicated source sentences.

Experiments on Tatoeba for all architectures are run for 50 epochs, with a~batch size of 128 and gradient clipping norm set to $10.0$. Encoder-decoder and RLST have weight decay set to $10^{-5}$, while Transformer has weight decay set to $10^{-4}$. Source and target tokens are converted to trainable vectors of length 256 initialized with $\mathcal{N}(0,\,1)$. There is a dropout applied to them with a~probability of $0.2$. We use Adam optimizer with default parameters and a constant learning rate equal to 0.0003. The reference encoder-decoder is presented in \cite{2015bahdanau+2}. Its encoder is a bidirectional GRU recurrent layer with 256 hidden neurons followed by a linear attention layer with 64 neurons. The decoder is a GRU recurrent layer with 256 hidden neurons followed by a dropout with 0.5 probability and a linear output layer with a number of neurons equal to the target's vocabulary size. The teacher forcing ratio for encoder-decoder during training is set to $1.0$. For the Transformer, we use the following parameters: the number of expected features in the encoder and decoder inputs is $256$, the number of heads in multiattention is $8$, the number of encoder and decoder layers is $6$, the dimension of feedforward layers is $512$, the dropout probability is $0.25$ and the teacher forcing ratio to $1.0$. For RLST, we use the following approximator. Input and previous output embeddings with a dimension of 256 are passed to a dense layer with 512 neurons, Leaky ReLU activation with negative slope set to 0.01 and dropout probability of 0.2. Its output is processed by four GRU layers with the hidden dimension of 512 and residual connections between them. The output of the last recurrent layer is passed to a dense layer with 512 neurons, Leaky ReLU activation with a negative slope set to 0.01, and a dropout probability of 0.5.
The output of the last dense layer is passed to the output linear layer with number of neurons equal to the target's vocabulary size and additional 2 neurons representing Q-values of actions. We also set $\gamma = 0.9$, $\varepsilon = 0.3$, $M = 3$, $N = 50000$, $\eta_{min} = 0.02$, $\eta_{max} = 0.2$, $\rho = 0.99$ and teacher forcing ratio to $1.0$. 

\subsection{IWSLT2014}

We conduct experiments on IWSLT2014 German-English and English-German datasets using the \textit{fairseq} framework \cite{ott2019fairseq}. Data is preprocessed using the script provided by the benchmark, which utilizes byte-pair encoding (BPE) \cite{sennrich2015neural}.
For every architecture, the training lasts for $100$ epochs with varying batch sizes to ensure that the maximum number of tokens in a batch equals $4096$ and the gradient clipping norm is set to $10.0$. The encoder-decoder and RLST architectures are trained using Adam optimizer with default parameters and constant learning rate scheduling with weight decay of $10^{-5}$. The Transformer is also trained using Adam optimizer, with parameters and a~learning rate scheduler described in \cite{2017vaswani+7}. For the encoder-decoder and the Transformer, we use the  \textit{lstm\_wiseman\_iwslt\_de\_en} architecture (based on \cite{wiseman2016sequence}) and \textit{transformer\_iwslt\_de\_en} (based on \cite{2017vaswani+7} with some changes), respectively. The encoder-decoder model has trainable source and target embeddings dimensions of $256$ without dropout. Its encoder is an LSTM layer with $256$ hidden neurons, and its decoder was also an LSTM layer with $256$ hidden neurons followed by an output layer with the number of neurons equal to the target's vocabulary size. The decoder uses the attention mechanism. The encoder and decoder layers have a dropout probability of $0.1$. The Transformer has the following parameters: The trainable source and target embeddings dimensions are $512$ without dropout, the number of neurons in feedforward layers is $1024$, the number of multiattention heads is $4$, the number of encoder and decoder layers is $6$ and a dropout probability of $0.1$. For RLST, we set trainable source and target embedding dimensions to $256$ with a dropout of 0.2 probability. In the case of IWSLT-En-De, we use the same approximator as in Tatoeba. For IWSLT-De-En, we changed the dimensions of dense and GRU layers from 512 to 768. We also set $\gamma = 0.9$, $\varepsilon = 0.30$, $M = 7$, $N = 100 000$, $\rho = 0.99$, $\eta_{min} = 0.02$, $\eta_{max} = 0.2$ and teacher forcing ratio to $1.0$. For encoder-decoder and Transformer, we set beam search width to $1$ and teacher forcing ratio to $1.0$. 

\subsection{Results}

The results are presented in Table \ref{tab:nmt_results}. For each dataset and architecture, we show BLEU values computed on a~test split from checkpoints for which the BLEU value on a~validation split was the highest. We also show the number of parameters for each model. The highest values of BLEU for each dataset are bolded. On the Tatoeba test confined to sentences of length up to 22 words, all three architectures achieved similar BLEU. However, in the test confined to longer sentences, RLST outperforms the other architectures in 4 language pairs out of 5, usually by a~large margin. The architecture to achieve the best results on fairseq datasets is the Transformer. RLST and the encoder-decoder with attention achieve similar BLEU on this benchmark. 

\begin{figure}
    \centering
    \includegraphics[width=1.0\linewidth]{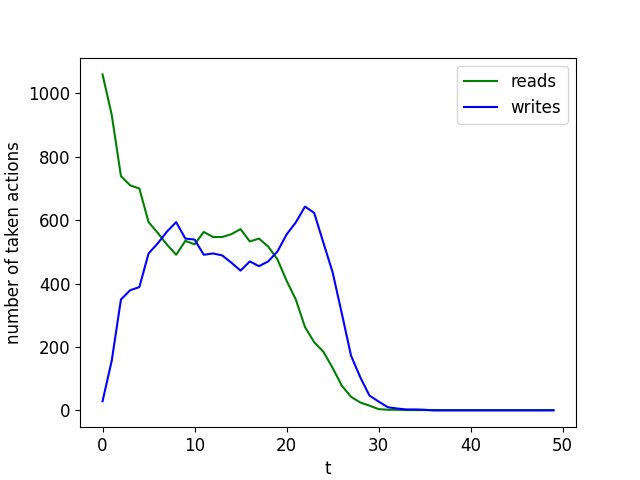}
    \caption{Processing of 1089 source sentences of length 15 from the Tat En-Ru test dataset by the RLST interpreter agent. The graph shows when the READ and WRITE actions are taken. }
    \label{fig:READWRITE}
\end{figure}

In order to gain an additional insight into the operation of the RLST interpreter agent we present in Figure~\ref{fig:READWRITE} the timing of taking the READ and WRITE actions. As one may expect, initially, the agent is mostly reading, then reading and writing ratios are roughly equal, and finally, the agent is mostly writing. It appears that the agent has read about five more words than it has written for most of the time. That seems to correspond to a~common intuition: A~human interpreter also needs to be delayed a~few words in producing an~accurate translation of a~speech.

\begin{table*}  
    \centering
    \begin{tabular}{|c|c c|c c|c c|}
    \hline
         Architecture $\rightarrow$ & \multicolumn{2}{c|}{Encoder-decoder} & \multicolumn{2}{c|}{Transformer} & \multicolumn{2}{c|}{RLST} \\
         Dataset $\downarrow$ & BLEU & Num. params & BLEU & Num. params & BLEU & Num. params \\
         \hline
         Tat Es-En & \textbf{50.33} & 16 637 504 & 50.19 & 15 906 560  & 50.02 &  17 122 050 \\
         Tat Es-En (L) & 16.52 & 16 637 504 & 13.42 & 15 906 560  & \textbf{20.57} & 17 122 050 \\
         Tat Fr-En & \textbf{53.95} & 18 170 504 & 53.89 & 16 597 832 & 53.05 & 18 093 898 \\
         Tat Fr-En (L) & 13.49 & 18 170 504 & 10.03 & 16 597 832 & \textbf{16.42} & 18 093 898 \\
         Tat En-Es & \textbf{45.14} & 20 248 794 & 44.63 & 16 647 066 & 45.09 & 18 819 484 \\
         Tat En-Es (L) & 16.1 & 20 248 794 & 12.39 & 16 647 066 & \textbf{ 21.07} & 18 819 484 \\
         Tat En-Ru & \textbf{47.71} & 32 271 740 & 47.11 & 21 664 316 & 47.37 & 26 171 966 \\
         Tat En-Ru (L) & 10.06 & 32 271 740 &  5.66 & 21 664 316 & \textbf{11.28} & 26 171 966 \\
         Tat En-De & \textbf{41.98} & 24 015 596 & 41.63 &  18 433 964 & 40.62 & 21 266 350 \\
         Tat En-De (L) & \textbf{10.95} & 24 015 596 & 9.5 &  18 433 964 & 10.2 & 21 266 350 \\
         IWSLT De-En & 24.13 & 7 178 728 & \textbf{32.17} & 42 864 640 & 23.28 & 24 223 210 \\
         IWSLT En-De & 19.01 & 7 748 240 & \textbf{26.13} & 43 999 232 & 18.32 & 15 331 986 \\
         \hline
    \end{tabular}
    \caption{BLEU scores on test splits and number of parameters for tested architectures. (L) on Tatoeba datasets denotes scores from long test split.}
    \label{tab:nmt_results}
\end{table*}

\section{Discussion} 
\label{sec:discussion} 

Our proposed interpreter agent RLST is designed to transform arbitrarily long sequences on-line. In each cycle of its operation, it performs the same number of computations in which it reads an input token or writes an output token. The agent has only limited memory space to store information about recently read tokens, a~context of these tokens defined by previous ones, and recently written tokens. Therefore, the agent is not a~method of choice for translating sentences of moderate length without any context. 

The RLST architecture outperformed others in the test on long sentences (longer than 22 words) taken from Tatoeba. The memory state of the interpreter agent preserved the context of the outputted words better than the attention mechanism managed to do in the reference architectures. We hypothesize that the sequential nature of human language makes it possible to translate properly separate parts of a~speech, but in order to do that, the end of each part must be identified. It appears that RLST manages to do it better in long sentences than the reference architectures. 

The~goal of a~large fraction of algorithms developed in computer science is to transform input data into output data whose size is unknown in advance. For some data types, it is natural to process them sequentially. These types include natural language, sound, video, and bioinformatic data, e.g., genetic. The experiments in Section~\ref{sec:experiments} confirm that our introduced RLST architecture is very well adapted to such data.

\section{Conclusions} 
\label{sec:conclusions} 

In this paper, we have presented the RLST architecture that transforms on-line sequences of arbitrary length without the need to define the trade-off between delay and quality. In the transformation process, it makes sequential decisions about whether to read an input token or write an output token. The architecture learns to make these decisions with reinforcement. The experimental study compared the architecture with state-of-the-art machine translation methods, namely the Transformer and the encoder-decoder with attention. Benchmark datasets taken from Tatoeba and IWSLT with seven language pairs were employed in the experiments. The RLST architecture solved a~more complex problem of on-line transformation than the reference methods, which produced output tokens knowing the entire source sequence. Even so, RLST produced translations of comparable quality. It also outperformed reference architectures in tests with long sentences (longer than 22 words) taken from Tatoeba. That confirms that it is particularly well suited to applications in which transformation of sequences of arbitrary lengths and/or on-line is required.



\end{document}